%% file: main.tex
\documentclass[10pt,twocolumn,letterpaper]{article}

\usepackage{iccv}              % To produce the CAMERA-READY version
\usepackage{multirow}
\usepackage{makecell}
\usepackage[normalem]{ulem}
\useunder{\uline}{\ul}{}
\usepackage{tabularx}  % 导入 tabularx 包
\usepackage{colortbl}
\usepackage{amsmath}


\definecolor{iccvblue}{rgb}{0.21,0.49,0.74}
\usepackage[pagebackref,breaklinks,colorlinks,allcolors=iccvblue]{hyperref}

\def\paperID{13196} % *** Enter the Paper ID here
\def\confName{ICCV}
\def\confYear{2025}

\title{FedPall: Prototype-based Adversarial and Collaborative Learning for Federated Learning with Feature Drift}

\author{Yong Zhang\textsuperscript{1, 3}, Feng Liang\textsuperscript{1}\thanks{Corresponding author.} , Guanghu Yuan\textsuperscript{1}, Min Yang\textsuperscript{2}, Chengming Li\textsuperscript{1} and Xiping Hu\textsuperscript{1, 3}\footnotemark[1]\\
Artificial Intelligence Research Institute, Shenzhen MSU-BIT University\textsuperscript{1}\\
Shenzhen Institutes of Advanced Technology, Chinese Academy of Sciences\textsuperscript{2}\\
School of Medical Technology, Beijing Institute of Technology\textsuperscript{3}\\
{\tt\small \{zhangyong2023@bit.edu.cn, fliang@smbu.edu.cn, yuangh@mail.ustc.edu.cn}\\
{\tt\small min.yang@siat.ac.cn, licm@smbu.edu.cn, huxp@smbu.edu.cn\}}
}

\begin{document}
\maketitle
\input{sec/0_abstract}    
\input{sec/1_introduction}

\input{sec/2_related_work}
\input{sec/3_method}
\input{sec/4_discussion}
\input{sec/5_experiments}
\input{sec/6_conclusion}
{
    \small
    \bibliographystyle{ieeenat_fullname}
    \bibliography{main}
}

\end{document}

%% file: sec/0_abstract.tex
\begin{abstract}
Federated learning (FL) enables collaborative training of a global model in the centralized server with data from multiple parties while preserving privacy. However, data heterogeneity can significantly degrade the performance of the global model when each party uses datasets from different sources to train a local model, thereby affecting personalized local models. Among various cases of data heterogeneity, feature drift, feature space difference among parties, is prevalent in real-life data but remains largely unexplored. Feature drift can distract feature extraction learning in clients and thus lead to poor feature extraction and classification performance. To tackle the problem of feature drift in FL, we propose FedPall, an FL framework that utilizes prototype-based adversarial learning to unify feature spaces and collaborative learning to reinforce class information within the features. Moreover, FedPall leverages mixed features generated from global prototypes and local features to enhance the global classifier with classification-relevant information from a global perspective. Evaluation results on three representative feature-drifted datasets demonstrate FedPall's consistently superior performance in classification with feature-drifted data in the FL scenario.\footnote{The code is available at \url{https://github.com/DistriAI/FedPall}.}
\end{abstract}

% \footnote{The code is attached in the supplementary material for review.}

%% file: sec/1_introduction.tex
\section{Introduction}
\label{sec:intro}

\begin{figure}[htbp]
    \centering
    \begin{subfigure}{0.6\linewidth}
        \includegraphics[height=0.8cm, width=\linewidth]{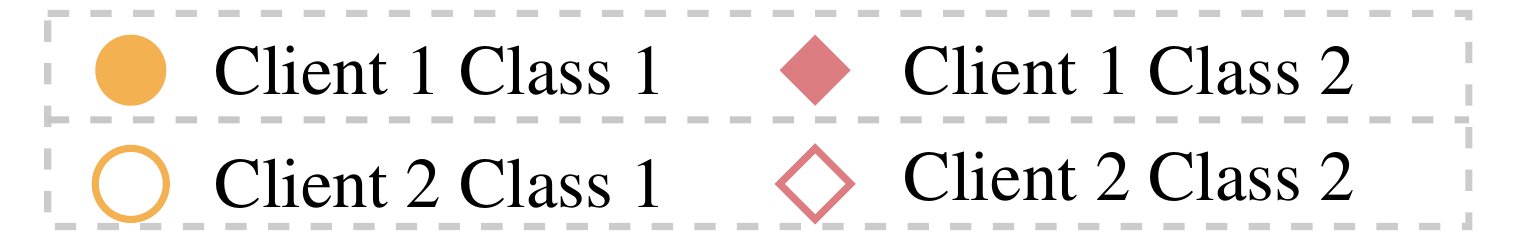}
    \end{subfigure}
    \begin{subfigure}{0.45\linewidth}
        \includegraphics[width=\linewidth]{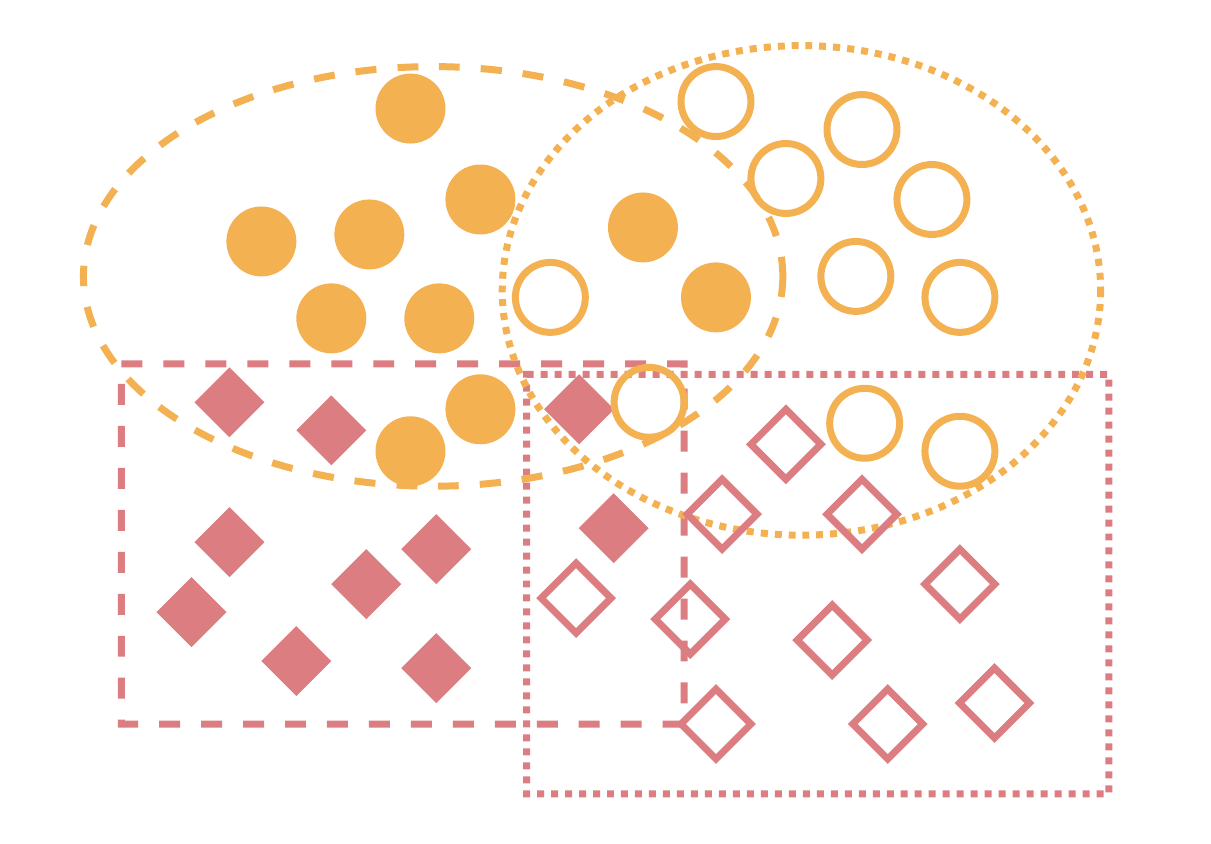}
        \caption{Feature drift}
        \label{fig:feature_a}
    \end{subfigure}
    \hfill
    \begin{subfigure}{0.45\linewidth}
        \includegraphics[width=\linewidth]{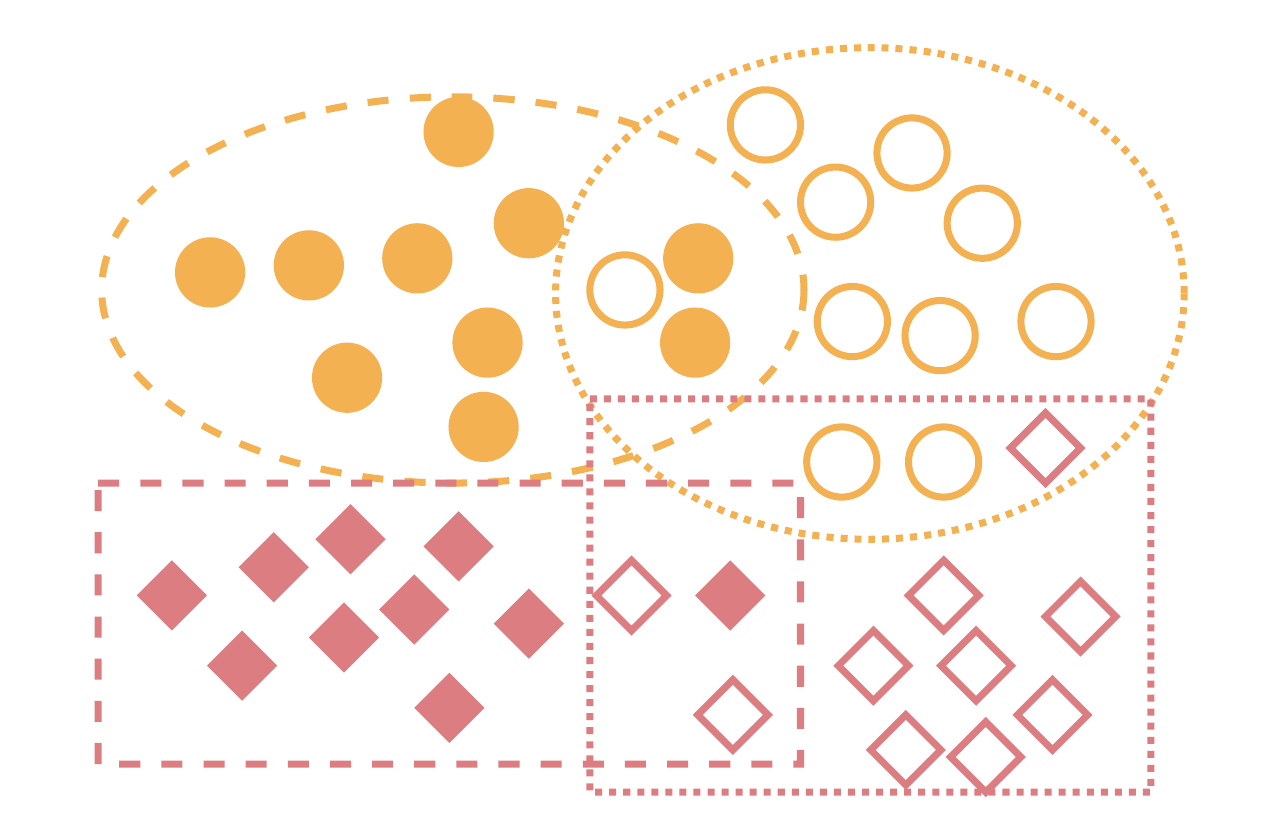}
        \caption{Traditional classification loss}
        \label{fig:feature_b}
    \end{subfigure}
    \begin{subfigure}{0.45\linewidth}
        \includegraphics[width=\linewidth]{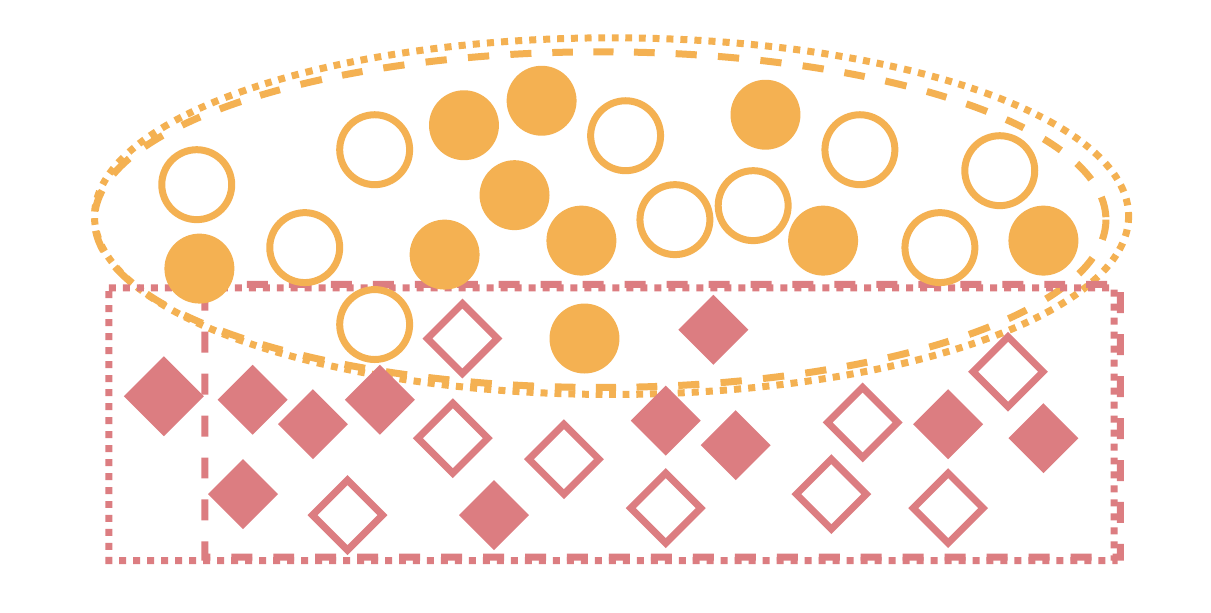}
        \caption{De-clientization}
        \label{fig:feature_c}
    \end{subfigure}
    \hfill
    \begin{subfigure}{0.45\linewidth}
        \includegraphics[width=\linewidth]{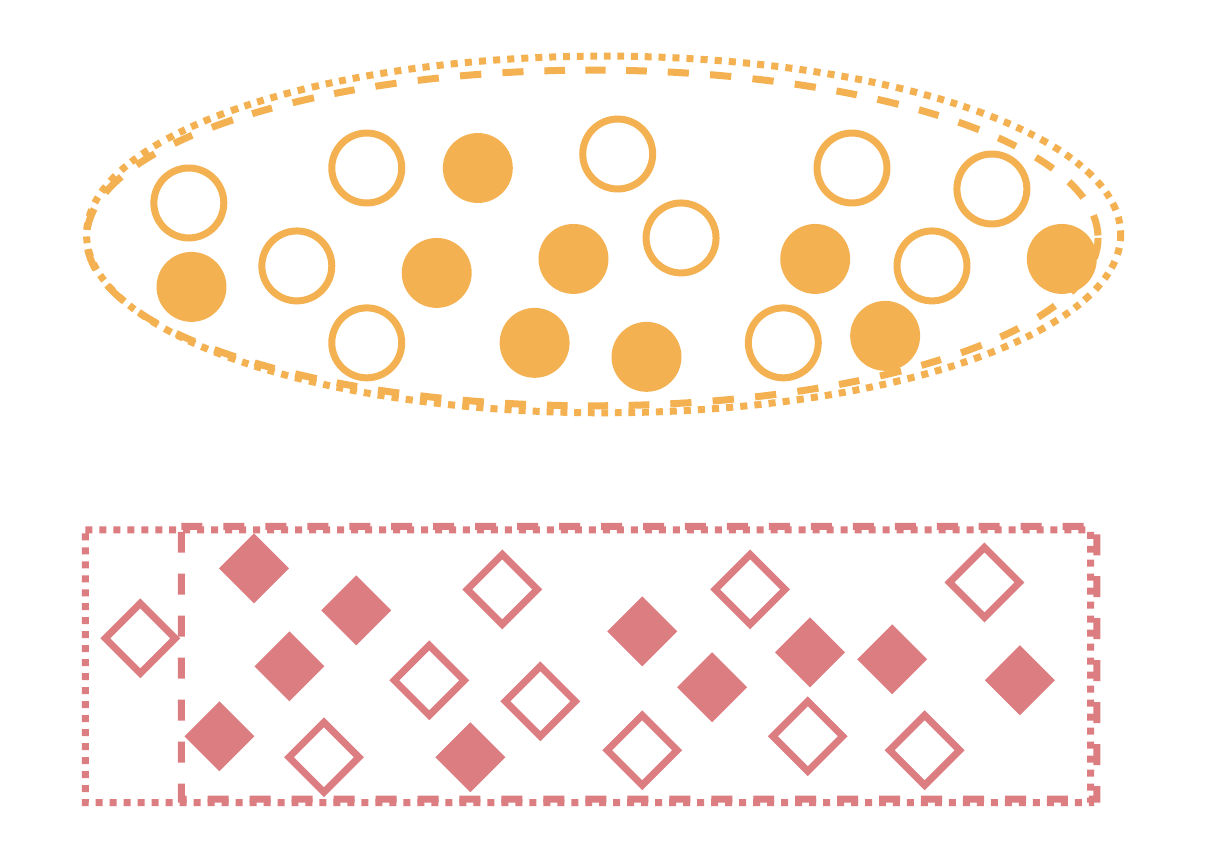}
        \caption{Prototype contrastive}
        \label{fig:feature_d}
    \end{subfigure}
    \caption{We show a schematic diagram of feature drift and use different techniques to drive feature distribution to update in different directions.}
    \label{fig:feature}
\end{figure}

Today, in computer vision, researchers often utilize large amounts of data from various parties to improve the accuracy of algorithms. However, this raises concerns, such as the potential for user privacy leakage caused by sharing private data~\cite{konevcny2015federated}. 
Federated learning (FL)~\cite{mcmahan2017communication} is proposed as a privacy-preserving distributed learning paradigm to address these challenges. 
In the FL paradigm, each party maintains a local client model and collaborates with others to train a global model on the server without sharing the original data, effectively protecting user privacy.

Particularly, the FL paradigm faces challenges from data heterogeneity~\cite{li2022new}. Due to the limited local view of each client, data distribution discrepancies arise across clients (commonly referred to as the non-independent and identically distributed data issue, or non-independent and identically distributed(non-IID) data issue), which can increase generalization error for local models and degrade the performance of the globally aggregated model~\cite{jiang2019improving}. Although recent work on non-IID data in FL primarily addresses issues such as stability, client drift, and heterogeneous label distributions among clients~\cite{li2020federated, karimireddy2019scaffold, zhao2018federated}, feature drift(i.e., variations in feature distributions across clients) is a prevalent yet underexplored challenge in FL. As shown in~\cref{fig:feature_a}, feature drift refers to the phenomenon where samples of the same class exhibit different feature distributions across different clients due to variations in data collection methods, devices, and other factors. This leads to ambiguous decision boundaries, severely impacting the classification performance of federated learning models. However, traditional classification losses (e.g., cross-entropy loss (CE loss), as shown in~\cref{fig:feature_b}) do not account for feature drift. As a result, the collaborative effect among different clients causes the feature space of the same-class samples to influence each other, leading to only slight or even no clearer distinction between decision boundaries for various classes within the same client. As illustrated in~\cref{fig:feature_c}, decentralization is a common approach to addressing feature drift, which involves ignoring inter-client differences to make the feature space of same-class samples across different clients more clustered. In feature drift scenarios, some state-of-the-art algorithms do not explore the role of loss functions but instead optimize local update algorithms to achieve decentralization. For example, FedBN~\cite{lifedbn} compresses the feature space across clients by adding batch normalization layers to local models. Some methods~\cite{zhu2021data, chen2023fraug} align feature spaces by sharing partial data for synthetic data augmentation. Additionally, ADCOL sends raw features to the server to update a amplifier and uses the Kullback-Leibler (KL) loss function to achieve decentralization. However, these methods have drawbacks: on the one hand, they may lead to loss of class information, and on the other hand, they pose potential risks of privacy leakage.

To tackle the above challenge, we propose \textit{FedPall}, a novel {\ul p}rototype-based {\ul a}dversarial and co{\ul ll}aborative learning framework for FL with feature drift. 
FedPall applies adversarial learning between clients and the server to unify feature spaces, as well as collaborative learning across clients to enhance global decision boundaries.
Specifically, FedPall uses adversarial learning to train a feature enhancer and uses KL divergence to align heterogeneous feature spaces between clients, while using prototype contrastive loss to reinforce class information (see~\cref{fig:feature_d}). Finally, adversarially aligned features are securely mixed with the global prototypes and uploaded to the server, where a global-view classifier is trained to enhance overall performance. Our contributions are summarized as follows:

\begin{itemize}
    \item We propose a novel FL framework, FedPall, to address the feature drift problem. FedPall introduces adversarial learning between clients and the server, and collaborative learning among clients aiming to project feature representations into a unified feature space and reinforce the intrinsic class information. This approach effectively mitigates the feature drift problem in FL settings.
    \item We develop a technical strategy that hierarchically integrates the global prototypes with local features to orchestrate client-server collaboration. The mixed prototype features are then used to train a global classifier, which induces the classifier to distill discriminative patterns through cross-client knowledge consolidation.
    \item Empirical evaluation on three typical feature-drifted benchmarks demonstrates that our proposed method achieves state-of-the-art classification accuracy. 
\end{itemize}

%% file: sec/2_related_work.tex
\section{Related Work}

In FL settings, the limited local view of each client directly induces the feature drift problem. Due to data-distribution differences, the same class label may have different feature representations, resulting in poor generalization of local models. Existing studies addressing this problem generally adopt two dominant paradigms: discriminative feature alignment and contrastive prototype learning.

Some studies have sought to address the problem of feature drift in FL by focusing on aligning feature representations. FRAug~\cite{chen2023fraug} employs data augmentation to generate synthetic embeddings encompassing global information and client-specific characteristics. FedSea~\cite{tan2023fedsea} aims to mitigate feature drift by aligning feature distributions to transform raw features into an IID format. FedCiR~\cite{li2024fedcir} addresses feature drift by maximizing mutual information between representations and labels while minimizing mutual information between client-specific representations conditioned on labels. Similarly, MOON~\cite{li2021model} encourages local models to align with the global feature distribution by constraining updates based on the similarity between local and global representations. Unlike traditional aggregation-based frameworks, ADCOL~\cite{li2023adversarial} employs adversarial learning to enforce a unified representation distribution across clients, thereby alleviating inter-client feature drift. However, it adopts a weak form of collaboration that does not address class boundaries from a global view. Moreover, its adversarial mechanism of directly transmitting features to the server introduces potential privacy risks. Our method adopts stronger collaborative learning to enhance global class boundaries and privacy-preserving adversarial learning to alleviate inter-client feature drift.
 
Some studies have focused on prototype-driven federated learning paradigms. Prototypes can compact feature embeddings through prototype abstraction, reducing communication bandwidth and preserving data privacy~\cite{wang2024federated}. Tan et al. \cite{tan2022fedproto, tan2022federated} proposed a supervised contrastive loss function leveraging both global and local prototypes to minimize the distance between feature representations and class prototypes, thereby addressing data heterogeneity. However, using the average feature as a prototype for each class may overlook intra-class variability within the feature space. To address this, MP-FedCL~\cite{qiao2023mp} utilizes clustering on the client side to generate multiple prototypes per class, capturing intra-class variation and mitigating feature drift arising from these differences. Following the effort of MP-FedCL, FedPLVM~\cite{wang2024taming} further enhances local training through a two-stage clustering process between clients and the server, incorporating an $\alpha$-sparsity prototype loss function to optimize performance. By leveraging the privacy-preserving nature of prototypes, this approach effectively addresses privacy and security concerns while using global prototypes to strengthen class-specific information within feature representations. In the FedPall framework, we enhance the collaboration between the client and the server through global prototypes. With the server's assistance, the client gains access to global category information, which helps to bring similar category data closer together and push data from different categories farther apart.
We also use mixed features with global category information to enhance the global classifier. 

%% file: sec/3_method.tex
\section{Method}

\begin{figure*}[htbp]
\centering
\begin{subfigure}{1\linewidth}
    \includegraphics[width=\linewidth]{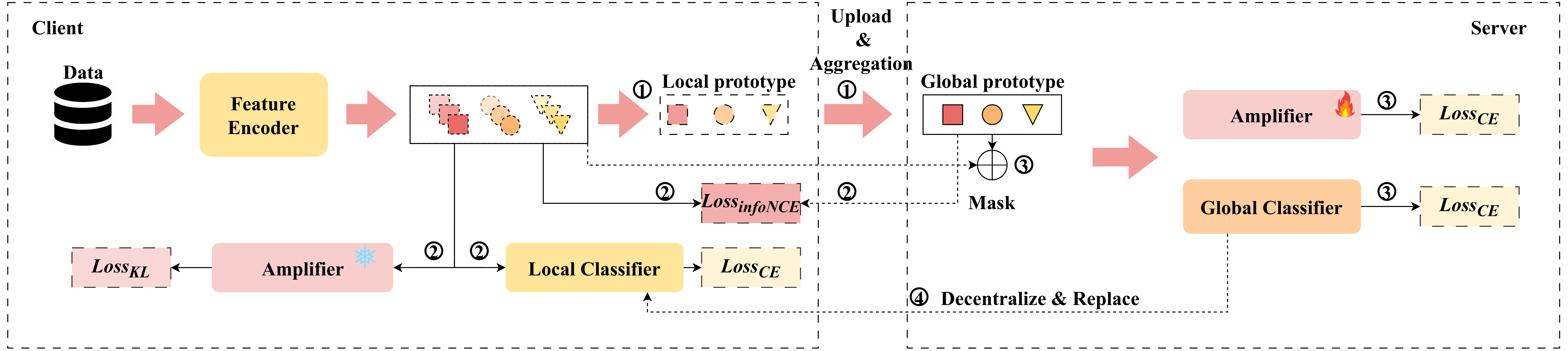}
    \caption{The overview of FedPall}
    \label{fig:flowchat}
\end{subfigure}
\begin{subfigure}{0.24\linewidth}
    \includegraphics[width=\linewidth]{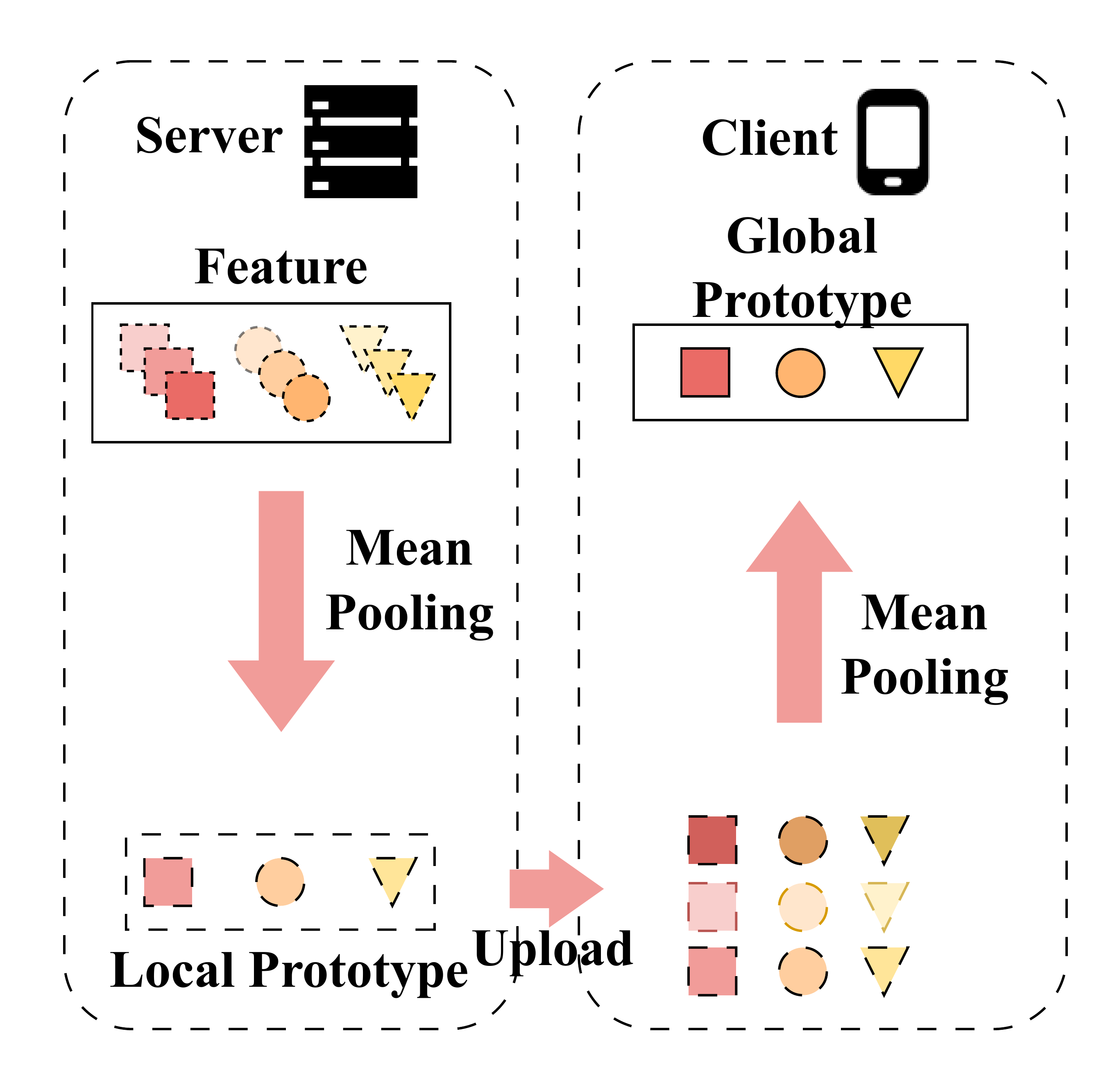}
    \caption{Generating Global Prototypes \textcircled{1}}
    \label{fig:flowchat1}
\end{subfigure}
\rule{0.5pt}{4.5cm}
\begin{subfigure}{0.24\linewidth}
    \includegraphics[width=\linewidth]{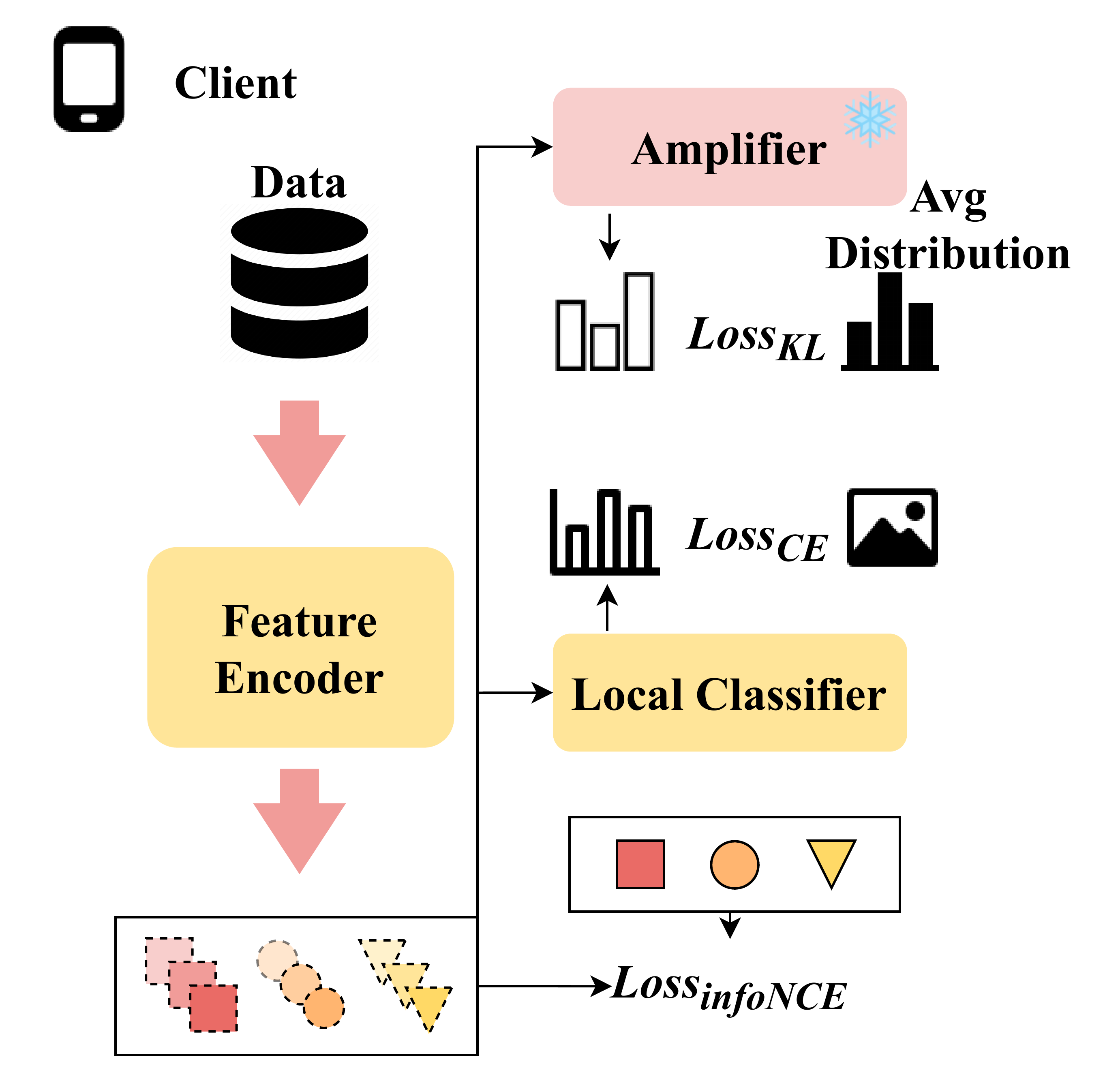}
    \caption{Training Local Models \textcircled{2}}
    \label{fig:flowchat2}
\end{subfigure}
\rule{0.5pt}{4.5cm}
\begin{subfigure}{0.23\linewidth}
    \includegraphics[width=\linewidth]{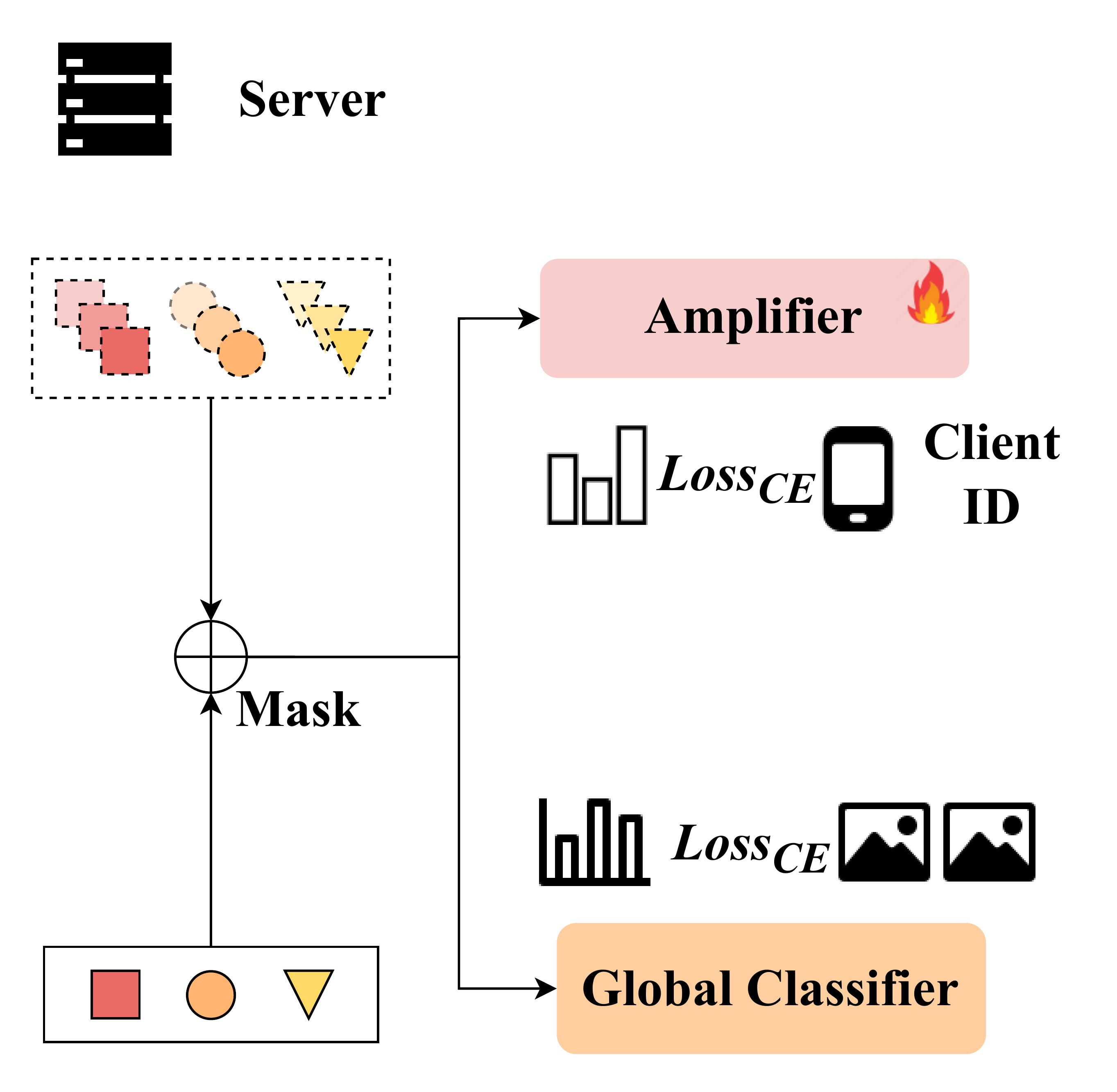}
    \caption{Training Global Model \textcircled{3}}
    \label{fig:flowchat3}
\end{subfigure}
\rule{0.5pt}{4.5cm}
\begin{subfigure}{0.25\linewidth}
    \includegraphics[width=\linewidth]{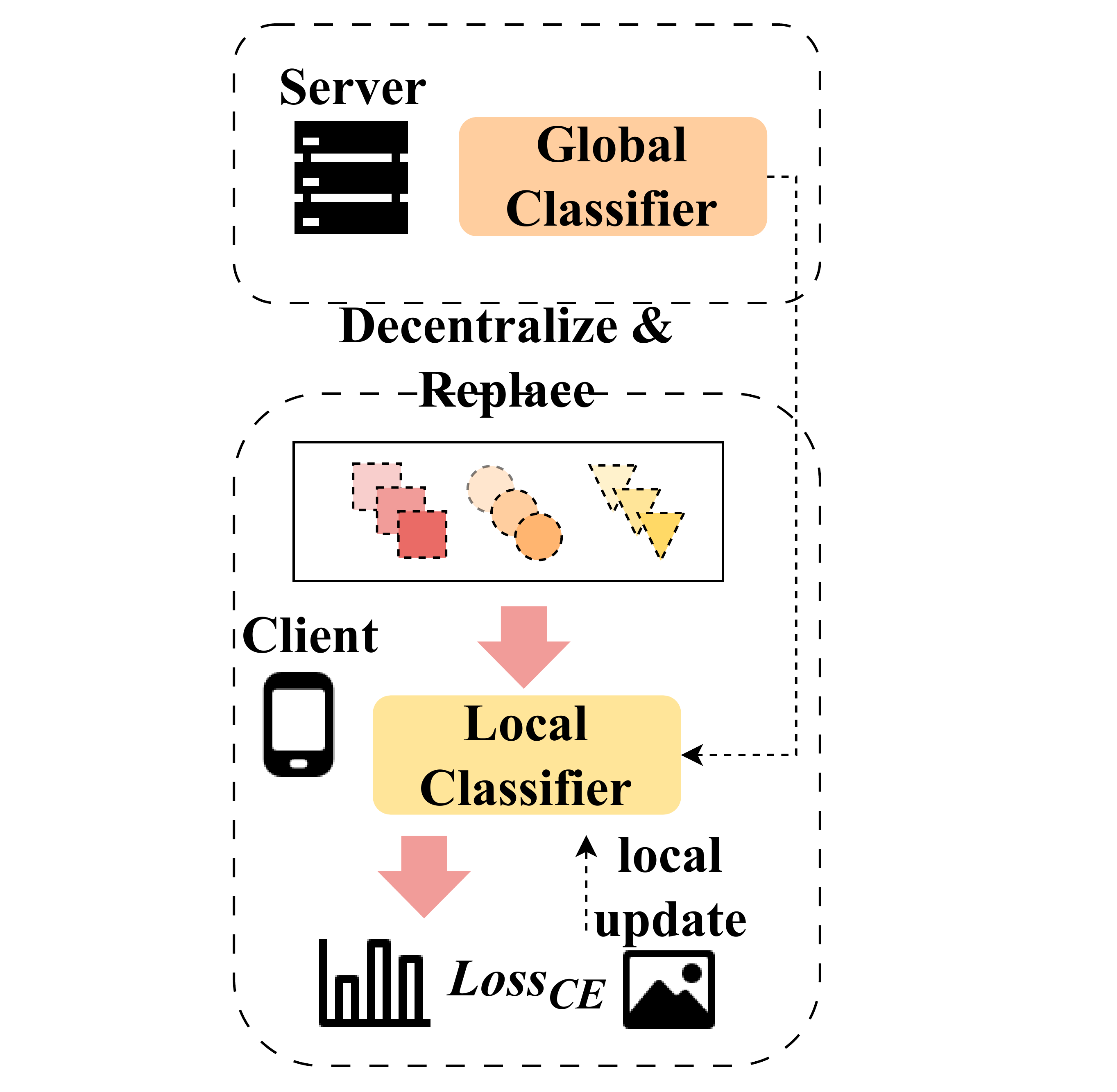}
    \caption{Decentralizing Global Classifier \textcircled{4}}
    \label{fig:flowchat4}
\end{subfigure}
\caption{The FedPall framework and detailed phases}
\label{fig:flowchatA}
\end{figure*}

\subsection{Problem Description}
We define \textit{feature drift} as follows: Given a dataset $\mathbb{D}$ with features $x$ and labels $y$, while the conditional distribution $P_i(x|y)$ differs across clients, the marginal distribution $P(y)$ remains the same. This means that the same label may have significantly different features across clients. For example, due to variations in environment, geographic location, and cultural differences, the structural features of houses can vary widely. 

% In the federated learning framework, there are a total of $N$ clients, each client $n$ has a private dataset $D_n$ In the setting of feature drift, the feature distribution $P(x)$ for each client is different, while the label distribution $P(y)$ is the same or similar~\cite{li2023adversarial}. Here, we present the overall optimization objective of the standard federated learning framework:
% \begin{equation}
%     \underset{w\in\mathbb{R}^{d_1}}{min}F(w):=\frac{1}{N}\sum_{n=1}^{N}f_n(w)
%     \label{eq1}
% \end{equation}
% where $w\in\mathbb{R}^{d_1}$ represents the parameter encoding of the global model, and $d_1$ denotes the dimensionality of the model parameters, and furthermore,
% \begin{equation}
%     f_n(w):=\mathbb{E}_{(x,y)\sim D_n}[f_n(w;x,y)]
%     \label{eq2}
% \end{equation}
% Where, $f_n(w)$ represents the expected loss obtained from client $n$ using the global model parameters under the dataset $D_n$.

With feature drift in FL, our goal is to optimize each client’s personalized model loss while leveraging the potential performance gains from collaborative learning across clients~\cite{zhou2023fedfa}. In the FedPall framework, there are a total of $N$ clients, each client $n$ has a private dataset $D_n$. Based on this goal, we formulate the overall optimization objective of the FedPall framework as follows:
\begin{equation}
    \underset{\theta_1, \theta_2,...,\theta_N\in\mathbb{R}^{d_1}}{min}F(\theta):=\frac{1}{N}\sum_{n=1}^{N}f_n(\theta_n),
    \label{eq3}
\end{equation}
where $f_n$ represents the expected loss obtained from client $n$ using the global model parameters under the dataset $D_n$, and $\theta_i$ represents the local model parameters of client $i$.

\subsection{FedPall Framework} 

Existing approaches to addressing the feature drift problem in FL typically focus on either collaborative learning or adversarial learning in isolation. This can result in models that either fail to adequately capture class-related information in the feature representations or exhibit persistent discrepancies in feature distributions across clients. To address these limitations, we propose integrating both adversarial and collaborative learning to effectively mitigate feature drift in FL settings. In this section, we present the FedPall framework by elaborating on its adversarial and collaborative learning. The framework of the overall approach is shown in \cref{fig:flowchat}. It is structured into four key procedures: generating global prototypes, training local models, training global model, and decentralizing global classifier. 

We define certain model symbols here that will be used in this section. 
The local model $F(\cdot)$ consists of two components, a feature extractor $G(\cdot)$ (e.g., Resnet50~\cite{he2016deep} for image data) and a classifier $H(\cdot)$. 
We use a multilayer perceptron (MLP) as our Amplifier, and except for the output layer, the number of nodes in other layers is consistent with that of the classifier.

\subsubsection{Generating Global Prototypes}
Several studies~\cite{mu2023fedproc, tan2022fedproto} suggest that category-centered prototypes are a privacy-friendly form of global knowledge. We leverage collaboration between clients to aggregate and generate global prototypes. Typically, the class prototype for each category is represented by the mean of the features for that category. The local prototype for the $k$-th category on client $n$ is defined as:
% \begin{equation}
%     c^k=\frac{1}{N^k}\sum_{(x,y)\in D^k}G(x)
%     \label{eq4}
% \end{equation}

\begin{equation}
    c_n^{k}=\frac{1}{N_n^k}\sum_{(x,y)\in D_n^k}G_n(x),
    \label{eq5}
\end{equation}
where $D_n^k$ and $N_n^k$ represent the data samples and the number of samples for the $k$-th category on client $n$, respectively. 

Gathering all the local prototypes together forms a local prototype set, which can be defined as:
\begin{equation}
    \mathcal{O}_n={\{c_n^1, c_n^2,...,c_n^K\}}\in \mathbb{R}^{K\times d},
    \label{eq:eq4}
\end{equation}
where $K$ represents the number of categories owned by each client, and $d$ denotes the output feature dimension.

Since we are only addressing the problem of feature drift, all clients have the same number of categories. Upon receiving the local prototype set and local label proportions $\mathcal{N}={\{\{N_n^k\}_{k=1}^{K}|n\in \mathcal{A}\}}$ sent by client set $\mathcal{A}$, the server integrates the local prototypes from all clients to form the global prototypes,
\begin{equation}
    \mathcal{G}^k=\sum_{n}\frac{N_n^k}{\sum_{n}{N_n^k}}c_n^k
    \label{eq:eq5}
\end{equation}

The global prototypes set can be represented as 
\begin{equation}
    \mathcal{G}=[\mathcal{G}^1,\ldots,\mathcal{G}^k,\ldots,\mathcal{G}^K]
    \label{eq:eq6}
\end{equation}

Next, the server sends the global prototype set $\mathcal{G}$ to each client to guide local model training. You can refer to~\cref{fig:flowchat1} for a better understanding of the generation of the global prototype.

\subsubsection{Training Local Models}
The goal of this module is to train an effective feature encoder that maps the raw data from different clients into a unified feature space, where the feature distributions are aligned and the class-related information is enhanced. 

As mentioned earlier, due to feature drift, training a local classifier alone is not sufficient for accurately classifying data with feature drift. To address this, we apply adversarial learning to train a feature encoder. Specifically, we use a global amplifier, trained on the server, which amplifies the heterogeneous information in the features from different clients. At the client's side, we apply the amplifier and use Kullback-Leibler (KL) divergence to reduce the heterogeneous information in the features, thereby creating an adversarial learning setup between the client and server. The objective is to improve the generalization ability of the feature encoders across clients while minimizing the client-specific heterogeneity in the feature representations. 
Let $x^{(i)}$ denote the $i$-th dimension of vector $x$, the KL divergence is calculated as:
\begin{equation}
    \begin{split}
        \mathcal{L}_{KL}&=\sum_{(\mathbf{x,y})\in D_{n}}D_{KL}(A_n(G(\mathbf{x}))||[\frac{1}{N}]^N)\\
        &=\sum_{(\mathbf{x},y)\in D_n}\sum_{i}^N A_n(G(\mathbf{x}))^{(i)}logNA_n(G(\mathbf{x}))^{(i)}.
    \end{split}
    \label{eq:KLloss}
\end{equation}
where, $x^{(i)}$ denotes the $i$-th dimension of vector $x$, $A_n(\cdot)$ represents the amplifier of the $n$-th client, $D_{KL}(P||Q)$ represents the KL divergence of P and Q.

After adversarial learning, although the feature representations of different clients are mapped to similar feature distributions, the class-related information may be blurred. To address this problem, we propose using contrastive learning to reinforce the class-related information within the feature encoder. To prevent information leakage, we enable collaboration between the global prototypes (server-side) and local features (client-side). Specifically, we employ the InfoNCE loss to minimize the distance between local features and their corresponding global prototypes, while maximizing the distance between local features and global prototypes of other classes. This approach strengthens the class-related representation within the feature encoder. The formula for the global-prototype contrastive loss is as follows:
\begin{equation}
    \mathcal{L}_{infoNCE}=\sum_{(\text{x},y)\in D_n}-log\frac{exp(sim(G(\text{x}),\mathcal{G}^y)/\tau)}{\sum_{y_\alpha\in A(y)}exp(sim(\text{x},\mathcal{G}^{(y_\alpha)})/\tau)},
    \label{eq8}
\end{equation}
where $A(y):=\{y_\alpha\in[1,|\mathcal{G}|]:y_\alpha\ne y\}$ is the set of labels distinct from $y$, $\tau$ is the temperature to adjust the tolerance for feature difference, and $sim(x,y)$ represents the cosine similarity of $x$ and $y$.

We combine adversarial learning and collaborative learning to address the feature drift problem in FL. By leveraging the two loss functions defined above, along with the local cross-entropy loss, we progressively train the local feature encoder at each client. The overall loss function for each client is as follows:
\begin{equation}
    \mathcal{L}={\mathcal{L}_{CE}}_{(\text{x},y)\sim D_n}(F(\text{x}),y)+\mu\mathcal{L}_{KL} + \delta\mathcal{L}_{infoNCE},
    \label{eq:eq9}
\end{equation}
where $\mu$ denotes the weight of the $\mathcal{L}_{KL}$ divergence and $\mathcal{L}_{CE}$ denotes the cross-entropy loss.

The local model $F(\cdot)$ is updated using \cref{eq:eq9}. For a comprehensive visualization of the global prototype generation process, consult~\cref{fig:flowchat2} where the workflow is systematically delineated.

\subsubsection{Training Global Model}
Due to the limited local view of clients, it is difficult for them to train an accurate classifier. Therefore, we upload the encrypted mixed features to the server to train a classifier with a global perspective. Additionally, we leverage the global view from the server to train an amplifier used for adversarial learning with the clients. 

For each feature  $z_n^{k}$ of class $k$ on client $n$, we obtain a prototype mixed feature by performing a weighted fusion with the corresponding global prototype:
% To avoid privacy leakage and further reduce the distance between local feature distributions and global feature distributions, we use global prototypes to encrypt the features. For each feature $z_n^{k}$ of class $k$ on client $n$, we perform a weighted fusion with the corresponding global prototype, as follows:
% Since the global prototypes are aggregated from normalized local prototypes, we first normalize the local features to balance the scales between the two.
\begin{equation}
    r_n^{k}=\alpha\times z_n^{k}+(1-\alpha)\times \mathcal{G}^k,
    \label{eq10}
\end{equation}
where $z_n^{k}\in Z_n^k=\{z_n^{i,k}\}_{i=1}^S$. $\alpha \sim U(u_f,u_r)$ are the mix parameters, with $u_f$ and $u_d$ representing two hyperparameters of a uniform distribution. $S$ represents the number of samples of the $k$-th category in client $n$.

Building on this, we employ a Bernoulli mask to further reduce the risk of privacy leakage,
\begin{equation}
\begin{split}
    Mask &= \{X_1, X_2, \ldots, X_d\}, \\
    X_i &\sim \text{Bernoulli}\left(\beta\right), \quad \forall i \in [1, d],
\end{split}
\label{eq11}
\end{equation}
The final output of the prototype mixing mechanism is derived by selecting the mixed feature elements based on the noise mask:
\begin{equation}
\tilde{r}_n^{k} = Mask\odot r_n^{k},
\label{eq12}
\end{equation}
where $\odot$ is an element-wise \textit{and} operator. 

After generating the prototype mixed features, the client will form the set $\mathcal{D}_{R_L}(R,Y)$ using the prototype mixed feature set $R_n=\{\tilde{r}_n^{1}, ...,\tilde{r}_n^{k},...,\tilde{r}_n^{K}\}\in\mathbb{R}^{K\times d}$ and the corresponding label set $Y$, which will be sent to the server.

The server updates the global amplifier $A$ using the mixed feature sets from each client along with the corresponding client IDs. Specifically, we first construct the dataset for training the amplifier, denoted as $\mathcal{D}_{R_I}(R,I)$, where $I$ represents the client IDs. The amplifier is then updated by minimizing the empirical risk:
\begin{equation}
    \mathbb{E}_{(R,I)\sim\mathcal{D}_{R_I}}\ell_{CE}(R,I).
\end{equation}

At the same time, the server updates the global classifier $C_g$ using the mixed prototype features and the class labels from the clients $Y$. This is done by minimizing the empirical risk:
\begin{equation}
    \mathbb{E}_{(R,Y)\sim\mathcal{D}_{R_L}}\ell_{CE}(R,Y).
\end{equation}

We show the training process of the global classifier in~\cref{fig:feature_c}.

\subsubsection{Decentralizing Global Classifier}
Finally, we deploy the global classifier $C_g$ to each client to replace the original local classifiers $C_c$. The purpose of this is to obtain a more generalized classifier that can alleviate the feature drift problem. To allow the global classifier to adapt to the personalized characteristics of local data, we retrain it on the client’s local data, thereby enhancing the classifier's accuracy and improving its performance on individual client data distributions. And you can understand the deployment method of the server-side global classifier through~\cref{fig:flowchat4}.

%% file: sec/4_discussion.tex
\section{Discussion}

\paragraph{Computational Cost}Compared to the standard federated learning model, our adversarial collaborative learning approach introduces an amplifier and a global classifier. However, the number of parameters for these two components is much smaller than those of the other components. In our design, both the amplifier and the classifier are designed as three-layer MLPs. Compared to the feature extractor (with a total of 23.5M parameters), the classifier (with a total of 1.32M parameters) and the amplifier (with a total of 1.33M parameters) account for approximately 5.61\% and 5.59\%, respectively. Moreover, the client-side amplifier remains frozen, functioning exclusively in the forward pass for loss calculation while being disabled during backpropagation cycles.

% \paragraph{Communication Efficiency}Compared to traditional federated learning methods like FedAvg, which sends local model parameters to the server, we use prototype-mixed features as the communication medium. Assume consistent tensor representations between model parameters and prototype-fused features, with the local model's feature extractor specifically employing a ResNet-50 architecture that produces 2048-dimensional embeddings. During the uploading process, our approach requires all clients to upload an average of 12,122 prototype-mixed features. In contrast, with FedAvg, each client needs to upload approximately 24.8 million parameters of model parameters. During the downloading process, our method only uses the parameters of the amplifier and global classifier as the communication content, which accounts for about 10.6\% of FedAvg's communication cost (refer to the above paragraph).

\paragraph{Communication Efficiency}Unlike traditional federated learning methods such as FedAvg, which transmit full local model parameters, our approach uses prototype-mixed features as the communication medium. Assuming consistent tensor representations between model parameters and prototype features, each client employs a ResNet-50-based feature extractor that generates 2048-dimensional embeddings. During upload, clients transmit an average of 12,122 prototype-mixed features, compared to approximately 24.8 million model parameters in FedAvg. For downloading, only the amplifier and global classifier parameters are transmitted, reducing communication cost to about 10.6\% of that in FedAvg (see previous paragraph).

\paragraph{Limitation} While our framework currently specializes in image recognition tasks, its extension to NLP or time-series analysis remains unexplored. Successful cross-domain adaptation requires two key developments: (1) establishing domain-specific feature representations and prototype definitions, and (2) redesigning loss functions according to task semantics. For NLP applications, this implies reconfiguring the standard classification paradigm into autoregressive prediction frameworks. Architectural adaptations are equally crucial - particularly the incorporation of RNN-based structures with inherent temporal modeling capabilities for sequential data processing.

%% file: sec/5_experiments.tex
\begin{table*}[!t]
\centering
\resizebox{\linewidth}{!}{
\begin{tabular}{cc|cccccccccc|c}
\hline
                                                          &                                      & SingleSet                           & FedAvg                              & FedProx                             & PerfedAvg                           & FedRep                              & FedBN                                     & MOON                                 & FedProto                            & ADCOL                                     & RUCR                                & ours(FedPall)                                         \\ \hline
\multicolumn{1}{c|}{}                                     & amazon                               & 73.96(2.71)                         & 56.94(2.46)                         & 56.60(2.57)                         & 57.12(2.17)                         & 45.31(1.88)                         & 40.80(15.75)                              & 51.74(16.11)                         & 69.44(2.10)                         & 73.26(4.37)                               & 52.08(8.53)                         & 72.92(1.38)                                  \\
\multicolumn{1}{c|}{}                                     & caltech                              & 44.74(3.15)                         & 46.52(4.63)                         & 50.96(5.19)                         & 50.81(1.56)                         & 38.37(4.92)                         & 33.93(6.48)                               & 41.33(13.62)                         & 39.41(6.32)                         & 37.19(1.68)                               & 44.30(1.03)                         & 44.74(8.74)                                  \\
\multicolumn{1}{c|}{}                                     & dslr                                 & 60.22(6.72)                         & 30.11(4.93)                         & 33.33(10.37)                        & 31.18(4.93)                         & 34.41(4.93)                         & 38.71(3.23)                               & 24.73(1.86)                          & 65.59(4.93)                         & 76.34(4.93)                               & 30.11(6.72)                         & 77.42(3.23)                                  \\
\multicolumn{1}{c|}{}                                     & webcam                               & 71.26(2.63)                         & 37.93(6.22)                         & 43.68(7.18)                         & 47.13(7.77)                         & 55.75(2.63)                         & 30.46(6.05)                               & 33.33(12.71)                         & 71.26(4.34)                         & 71.26(2.63)                               & 37.36(4.98)                         & 74.71(1.00)                                  \\ \cline{2-13} 
\multicolumn{1}{c|}{\multirow{-5}{*}{\textbf{Office-10}}} & \cellcolor[HTML]{DAE8FC}\textbf{avg} & \cellcolor[HTML]{DAE8FC}62.54(0.38) & \cellcolor[HTML]{DAE8FC}42.88(1.18) & \cellcolor[HTML]{DAE8FC}46.14(2.64) & \cellcolor[HTML]{DAE8FC}46.56(2.89) & \cellcolor[HTML]{DAE8FC}43.46(1.34) & \cellcolor[HTML]{DAE8FC}35.97(6.54)       & \cellcolor[HTML]{DAE8FC}37.78(10.89) & \cellcolor[HTML]{DAE8FC}61.43(1.74) & \cellcolor[HTML]{DAE8FC}{\ul 64.51(1.79)} & \cellcolor[HTML]{DAE8FC}40.96(0.61) & \cellcolor[HTML]{DAE8FC}\textbf{67.45(2.69)} \\ \hline
\multicolumn{1}{c|}{}                                     & MNIST                                & 95.51(0.22)                         & 92.86(2.24)                         & 91.79(2.95)                         & 90.10(4.79)                         & 86.54(6.08)                         & 96.69(0.11)                               & 93.41(1.14)                          & 96.37(0.50)                         & 96.30(0.41)                               & 92.59(1.96)                         & 97.24(0.42)                                  \\
\multicolumn{1}{c|}{}                                     & SVHN                                 & 71.09(0.91)                         & 77.39(0.21)                         & 76.92(0.28)                         & 75.64(0.42)                         & 67.17(1.73)                         & 79.44(0.25)                               & 79.63(0.75)                          & 72.50(0.29)                         & 75.12(2.08)                               & 77.94(0.25)                         & 78.00(0.36)                                  \\
\multicolumn{1}{c|}{}                                     & USPS                                 & 86.40(0.27)                         & 89.25(0.89)                         & 89.23(1.41)                         & 88.69(0.69)                         & 89.95(2.95)                         & 90.07(0.54)                               & 81.76(0.72)                          & 87.01(0.83)                         & 86.72(1.25)                               & 88.85(2.39)                         & 87.28(1.29)                                  \\
\multicolumn{1}{c|}{}                                     & SynthDigits                          & 95.15(0.13)                         & 95.49(0.07)                         & 95.39(0.12)                         & 95.00(0.16)                         & 94.21(0.78)                         & 95.61(0.06)                               & 96.63(0.23)                          & 95.29(0.61)                         & 96.43(0.29)                               & 95.98(0.17)                         & 95.26(0.43)                                  \\
\multicolumn{1}{c|}{}                                     & MNIST-M                              & 76.56(0.40)                         & 73.81(1.45)                         & 74.02(1.49)                         & 73.21(0.78)                         & 69.11(0.94)                         & 76.25(0.39)                               & 72.16(0.92)                          & 78.27(1.20)                         & 78.28(4.39)                               & 72.65(0.37)                         & 85.90(1.39)                                  \\ \cline{2-13} 
\multicolumn{1}{c|}{\multirow{-6}{*}{\textbf{Digits}}}    & \cellcolor[HTML]{DAE8FC}\textbf{avg} & \cellcolor[HTML]{DAE8FC}84.94(0.06) & \cellcolor[HTML]{DAE8FC}85.76(0.86) & \cellcolor[HTML]{DAE8FC}85.47(1.14) & \cellcolor[HTML]{DAE8FC}84.53(1.31) & \cellcolor[HTML]{DAE8FC}81.40(2.47) & \cellcolor[HTML]{DAE8FC}{\ul 87.61(0.11)} & \cellcolor[HTML]{DAE8FC}84.72(0.60)  & \cellcolor[HTML]{DAE8FC}85.89(0.23) & \cellcolor[HTML]{DAE8FC}86.57(1.32)       & \cellcolor[HTML]{DAE8FC}85.60(0.87) & \cellcolor[HTML]{DAE8FC}\textbf{88.74(0.15)} \\ \hline
\multicolumn{1}{c|}{}                                     & art\_painting                        & 33.58(0.84)                         & 25.79(1.93)                         & 24.33(4.14)                         & 26.52(2.19)                         & 26.93(3.32)                         & 36.66(1.76)                               & 30.58(1.97)                          & 32.68(0.70)                         & 34.87(1.15)                               & 24.66(1.10)                         & 35.60(0.56)                                  \\
\multicolumn{1}{c|}{}                                     & cartoon                              & 58.53(2.48)                         & 45.36(2.29)                         & 51.38(0.56)                         & 48.27(1.24)                         & 44.37(2.09)                         & 55.63(1.95)                               & 51.52(1.78)                          & 57.25(1.51)                         & 57.18(0.80)                               & 47.49(3.32)                         & 59.73(2.34)                                  \\
\multicolumn{1}{c|}{}                                     & photo                                & 63.01(1.93)                         & 48.66(3.08)                         & 49.55(1.95)                         & 46.88(2.64)                         & 41.94(2.76)                         & 66.07(1.04)                               & 53.02(3.34)                          & 64.00(1.34)                         & 62.12(1.98)                               & 47.48(6.22)                         & 64.69(1.29)                                  \\
\multicolumn{1}{c|}{}                                     & sketch                               & 79.70(0.13)                         & 49.03(1.98)                         & 40.74(1.54)                         & 44.42(3.77)                         & 40.48(1.25)                         & 79.57(1.65)                               & 55.12(1.37)                          & 79.61(0.81)                         & 80.12(1.03)                               & 42.17(2.40)                         & 82.23(0.71)                                  \\ \cline{2-13} 
\multicolumn{1}{c|}{\multirow{-5}{*}{\textbf{PACS}}}      & \cellcolor[HTML]{DAE8FC}\textbf{avg} & \cellcolor[HTML]{DAE8FC}58.70(1.23) & \cellcolor[HTML]{DAE8FC}42.21(1.59) & \cellcolor[HTML]{DAE8FC}41.50(1.82) & \cellcolor[HTML]{DAE8FC}41.52(1.90) & \cellcolor[HTML]{DAE8FC}38.43(1.11) & \cellcolor[HTML]{DAE8FC}{\ul 59.48(1.44)} & \cellcolor[HTML]{DAE8FC}47.56(0.88)  & \cellcolor[HTML]{DAE8FC}58.39(0.25) & \cellcolor[HTML]{DAE8FC}58.57(0.58)       & \cellcolor[HTML]{DAE8FC}40.45(1.98) & \cellcolor[HTML]{DAE8FC}\textbf{60.56(0.36)} \\ \hline
\end{tabular}
}

\caption{The top-1 accuracy (\%) of each algorithm on each sub-dataset of the Office-10, Digits, and PACS datasets is compared, along with the average top-1 accuracy across all sub-datasets. The mean and standard deviation (std) from three random trials (using different random seeds, with other experimental settings remaining the same) are reported. The highest accuracy for each dataset is highlighted in bold, and the second-highest accuracy is underlined.}
\label{tab:ResultInOffice}
\end{table*}

\section{Experiments}
\subsection{Experimental Setup}
\paragraph{Datasets}
We conduct experiments on three publicly available feature drift datasets: Digits~\cite{zhou2020learning}, Office-10~\cite{gong2012geodesic}, and PACS~\cite{li2017deeper}. Specifically, (1) the Digits dataset consists of five different domain sources: MNIST~\cite{lecun1998gradient}, SCHN~\cite{netzer2011reading}, USPS~\cite{hull1994database}, SynthDigits~\cite{ganin2015unsupervised}, and MNIST-M~\cite{ganin2015unsupervised}; (2) the Office-10 dataset includes four distinct sources: Amazon, Caltech, DSLR, and WebCam; (3) the PACS dataset consists of four sources: Art Painting, Cartoon, Photo, and Sketch. Datasets Office-10 and PACS are real-world images from natural scenes, which inherently exhibit feature drift due to the diversity of their sources. 
Digits is a digit recognition dataset. In line with~\cite{lifedbn, tan2022federated}, we do not use the entire Digits dataset for feature transformation experiments but rather a subset of 10\% of the data. For datasets Office-10 and PACS, we used all of the datasets for the experiment. Additionally, we split each dataset into training and testing sets with an 8:2 ratio.

% We will conduct experiments using three datasets: Digits~\cite{zhou2020learning}, Office-10~\cite{gong2012geodesic}, and PACS~\cite{li2017deeper}. (1) Digits includes five data sources from different domains: MNIST~\cite{lecun1998gradient}, SCHN~\cite{netzer2011reading}, USPS~\cite{hull1994database}, SynthDigits~\cite{ganin2015unsupervised}, and MNIST-M~\cite{ganin2015unsupervised}. (2) Office-10 contains four different sources: Amazon, Caltech, DSLR, and WebCam. (3) PACS consists of four different sources: Art-painting, Cartoon, Photo, and Sketch. The second and third datasets are real-world datasets in natural scenes, and since they originate from different sources, they naturally form a feature non-IID setting. Similar to~\cite{li2021FedBN, tan2022federated}, we do not select all data from the Digits dataset for feature drift experiments but instead use 10\% of the data. For Office-10 and PACS, we select all data for the experiments. And we divide each dataset into training and test sets using an 8:2 split.

\paragraph{Baselines}
We compare FedPall with ten baselines, including SingSet (where each client independently trains a model).
FedAvg~\cite{mcmahan2017communication} is the most classic federated learning algorithm, while FedProx~\cite{li2020federated}, PerFedAvg~\cite{fallah2020personalized}, and FedRep~\cite{collins2021exploiting} are personalized federated learning methods.
FedBN~\cite{lifedbn}, ADCOL~\cite{li2023adversarial}, MOON~\cite{li2021model} and FedProto~\cite{tan2022fedproto} are personalized federated learning algorithms for cross-domain learning, all of which address the issue of non-IID features to some extent. In addition, we explored the ability of RUCR~\cite{huang2024federated} to solve the feature drift problem.

\paragraph{Model and Hyper-parameter Setup}

All algorithms adopt identical local model architectures for fair comparison. Each local model contains: (1) a ResNet-50 feature extractor (excluding classifier layer), (2) a three-layer MLP classifier with 512 hidden units, and (3) a three-layer MLP amplifier with 2048 input/512 hidden units. Output dimensions for classifier and amplifier are set according to dataset categories and data sources, respectively. We maintain client count equal to data sources (one source per client). Local training uses 5 epochs for Digits, and 10 epochs for Office-10 and PACS. Global training employs 100 epochs throughout. Optimization uses SGD (lr=0.01). Except for the digits dataset, for which the values of $\mu$ and $\delta$ are set to 0.7 and 0.3 respectively, the values of $\mu$ and $\delta$ are set to 0.1 for all other datasets. Common loss hyperparameters remain fixed across algorithms. All experiments run on NVIDIA RTX 4090 GPUs.

% By default, the number of clients equals the number of data sources in the corresponding dataset, and the output dimension of the predictor matches the number of classes in the dataset. On the Digits dataset, we set the number of local training rounds to 5. On the Office-10 and PACS datasets, we set the number of local training rounds to 10. The global training rounds for all datasets are set to 10. We use the SGD optimizer with a learning rate of 0.01 for training. Our experiments are conducted on an NVIDIA GeForce RTX 4090.
\subsection{Experimental Analysis}

We conduct the evaluation on three publicly available feature-drifted datasets (Digits, Office-10, and PACS) and compare the performance of the FedPall framework with classical and state-of-the-art baselines. As shown in Table~\ref{tab:ResultInOffice}, our proposed framework achieves state-of-the-art accuracy on all three datasets.
% validating the effectiveness of the FedPall framework. 

We first discuss the experimental results based on each individual dataset.
On the Office-10 dataset, the overall accuracy of the FedPall framework surpasses that of the second-best method, ADCOL, by approximately 3 percentage points. 
% By examining the accuracy across the four sub-datasets (as for the four clients), FedPall improves the overall performance by significantly elevating the performance of some sub-datasets while not decreasing that of others.
On the Digits dataset, it is evident that FedPall outperforms all other models, achieving an accuracy that is approximately 1.1 percentage points higher than the second-best model, FedBN. 
% Moreover, FedPall surpasses the popular baseline with a comparative idea, ADCOL, by about 2.2 percentage points. 
The Digits dataset contains images that are relatively easy to classify, and the degree of feature drift is smaller compared to the Office-10 dataset. All baseline models achieve reasonably good accuracy on this dataset. Specifically, adversarial learning helps mitigate the heterogeneous information in the MNIST-M client. 
% In addition, collaborative learning allows the local model on the MNIST-M client to benefit from the knowledge shared by other clients and the server. 
Similarly, our algorithm demonstrates strong performance on the PACS dataset, achieving an overall accuracy that is approximately 1.1 percentage points higher than the second-highest result produced by FedBN. FedPall achieves the highest or second-highest accuracy across all sub-datasets. 
% Specifically, it achieves near-optimal results on the Art Painting and Photo sub-datasets, with accuracy only slightly lower (by no more than 2 percentage points) than the best-performing model. On the Cartoon and Sketch sub-datasets, FedPall outperforms all other methods, with accuracy higher by approximately 1.2 and 2.1 percentage points, respectively, compared to the second-best model. These results further validate that the FedPall framework effectively addresses the feature drift problem in federated learning settings.
% Our method not only performs well on simpler benchmark datasets (Digits) but also demonstrates strong performance in highly heterogeneous real-world scenarios (Office-10, PACS). This highlights the broad applicability and potential of our method across diverse settings.

We also discuss the performance of FedPall as compared to other state-of-the-art baselines across all three datasets.
The average accuracy of FedPall consistently outperforms that of ADCOL in all three datasets, achieving an increase ranging from about 1.1 to 2.9 percentage points.
In addition, even though FedBN can achieve accuracy comparable to our method on datasets Digits and PACS, our method outperforms it significantly by 31.5 percentage points on datasets Office-10.
As mentioned earlier, the Office-10 dataset comes from real-world data, where feature drift is particularly prominent, and there is also a significant distribution difference between the training and testing sets, leading to the suboptimal performance of the FedBN method on this dataset. In contrast, the special design incorporating both adversarial and collaborative learning in FedPall enables it to adapt well to the Office-10 dataset.

\subsection{Ablation Study}
\paragraph{Effect of loss combination}

\begin{figure}[t]
    \centering
    \includegraphics[width=.9\linewidth]{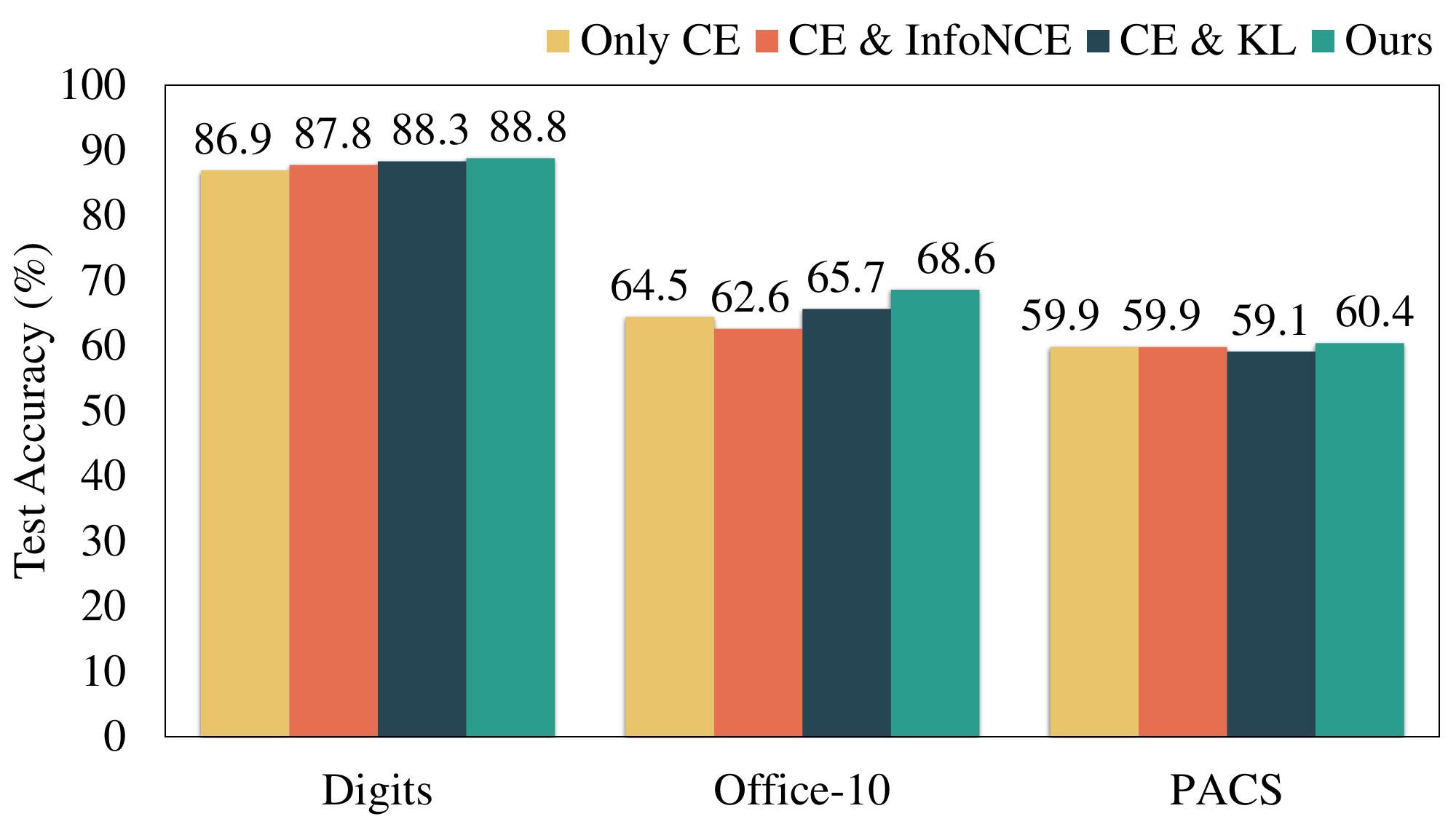}
    \caption{We evaluate the top-1 accuracy averaged over all clients using different loss function combinations on different datasets.}
    \label{fig:ablation-office}
\end{figure}
In this section, we analyze the impact of KL loss and InfoNCE loss on the performance of the local feature encoder. We conduct ablation studies with three configurations: (1) removing both KL and InfoNCE losses, (2) removing only KL loss, and (3) removing only InfoNCE loss. As shown in Figure \ref{fig:ablation-office}, the algorithm performs best when all losses are retained, which validates the reliability of the loss combination we designed. 
% Specifically, on the Office-10 dataset, our method outperforms the model using only CE loss by nearly 4 percentage points. Interestingly, the combination of CE loss and InfoNCE loss performs even worse than using only CE loss. This suggests that, when the feature drift problem is more pronounced, reinforcing the category information solely through InfoNCE loss may exacerbate the feature drift. Although the combination of CE loss and KL loss performs better than using CE loss alone on the Office-10 dataset, its performance on the PACS dataset falls below expectations (even lower than CE loss), indicating that the model's robustness is weaker when using CE and KL losses together. This can lead to some loss of category information, and while the model can adapt to some heterogeneous datasets, it remains highly unstable. 

Specifically, on the Office-10 dataset, our method outperforms the model trained with only CE loss by nearly 4 percentage points. Notably, combining CE loss with InfoNCE loss yields worse results than using CE loss alone, suggesting that in the presence of severe feature drift, reinforcing category information through InfoNCE may amplify the drift. While the CE + KL loss combination performs better than CE loss alone on Office-10, it underperforms on the PACS dataset—even falling below CE loss—indicating reduced robustness. This suggests that CE + KL may compromise category information, leading to instability across heterogeneous datasets.

\begin{figure}[t]
    \centering
    \includegraphics[width=0.9\linewidth]{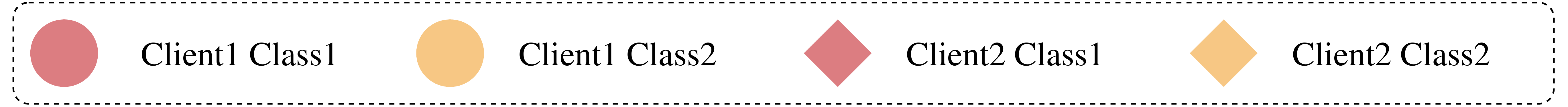}
    \vspace{10mm}
    \includegraphics[width=1\linewidth]{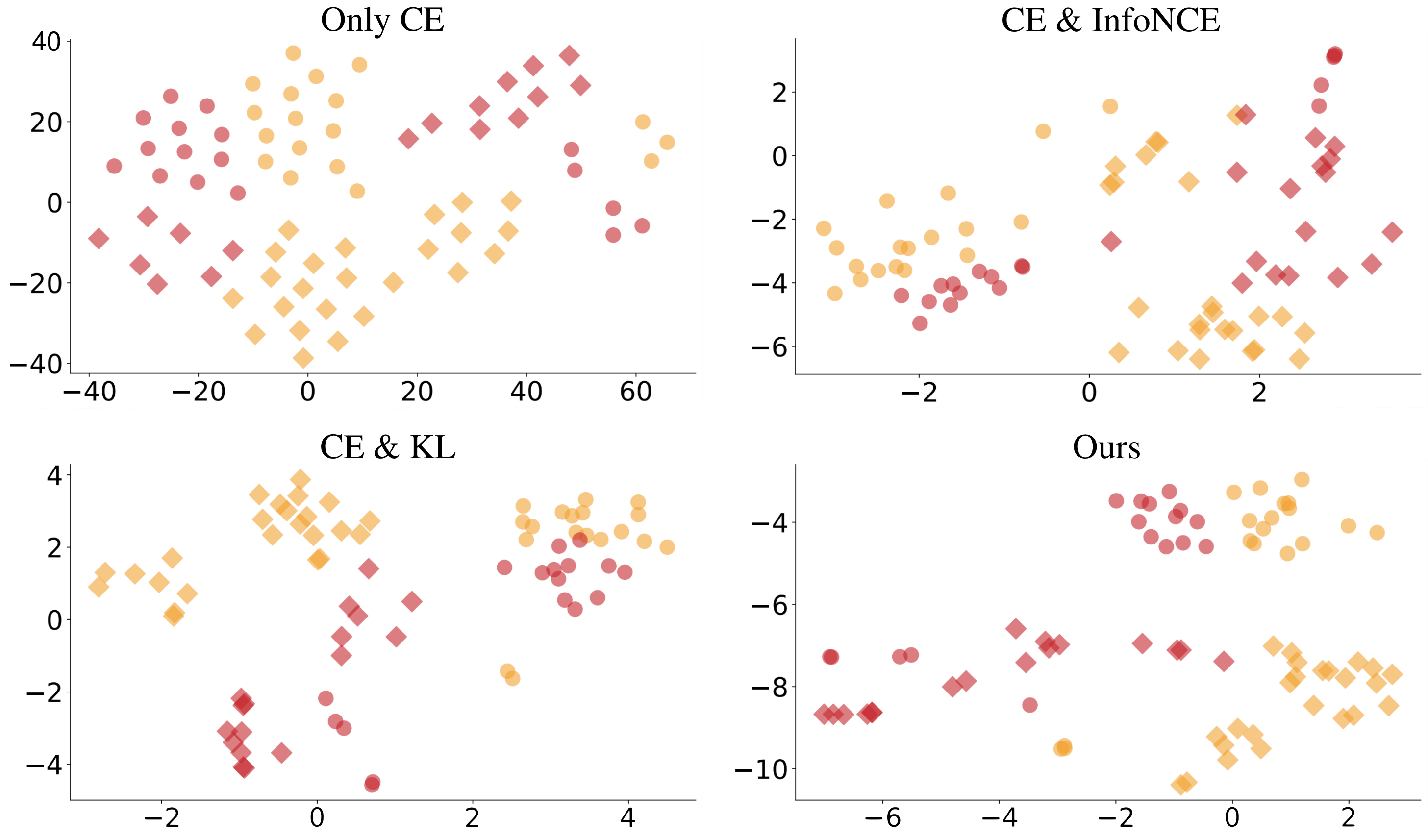}
    \caption{We plotted the feature distribution of different categories under different clients, corresponding to the four loss combination strategies of \cref{fig:ablation-office}}
    \label{fig:scatter}
\end{figure}

To assess the impact of KL and InfoNCE losses on feature distributions, we visualize class-wise features across clients in the Office-10 dataset using t-SNE projections of randomly sampled data points (Figure~\ref{fig:scatter}). The CE-only model poorly mitigates feature drift, with Client 1 showing unclear decision boundaries. Adding InfoNCE improves intra-client class separation but fails to resolve inter-client drift, leading to ambiguous global boundaries. The CE+KL combination reduces cross-client distances for the same class, yielding clearer global boundaries; however, it compresses intra-class spacing in Client 1, causing overlapping clusters and outliers that hinder local classification. In contrast, our unified loss balances these effects: KL aligns same-class features across clients, while CE and InfoNCE promote intra-client separation. This coordination produces compact, well-separated clusters, improving classification and validating our method’s effectiveness.
% In summary, our method not only outperforms other combinations in the simple handwritten digit recognition scenario but also demonstrates superior performance on other real-world datasets. This highlights the robustness and generalization capability of the loss combination we designed.

In summary, our method outperforms other loss combinations in the simple handwritten digit recognition task and achieves superior results on real-world datasets, demonstrating strong robustness and generalization.

% We compared the effects of different loss function combinations on our method, and the results are shown in \cref{fig:ablation-office}. It can be seen that our method performs better than other loss function combinations on different datasets, indicating that our method has good robustness. On the Office-10 dataset, our method is nearly 4 percentage points higher than the CE loss alone, while the combination of CE loss and contrast loss is even lower than the accuracy of CE loss alone. This shows that although the contrast loss function can deal with simple feature drift situations, when the dataset has serious data heterogeneity, the contrast loss is difficult to achieve effective classification.Although the combination of CE loss and KL loss performs better than the CE loss alone on the Office-10 dataset, its performance in the PACS dataset is lower than expected (lower than CE loss). This phenomenon reflects that the model obtained by CE loss and KL loss is weak in robustness. Although it can adapt to the situation of some heterogeneous datasets, it is extremely unstable.In short, our method is not only higher than other combinations in simple handwritten digit recognition scenarios, but also has excellent performance on all real datasets, which shows that the model we trained has strong robustness and generalization ability.

\paragraph{Hyperparameter sensitivity analysis.}
\begin{figure}[t]
  \centering
  \begin{subfigure}[b]{0.8\linewidth}
    \centering
    \includegraphics[width=\linewidth]{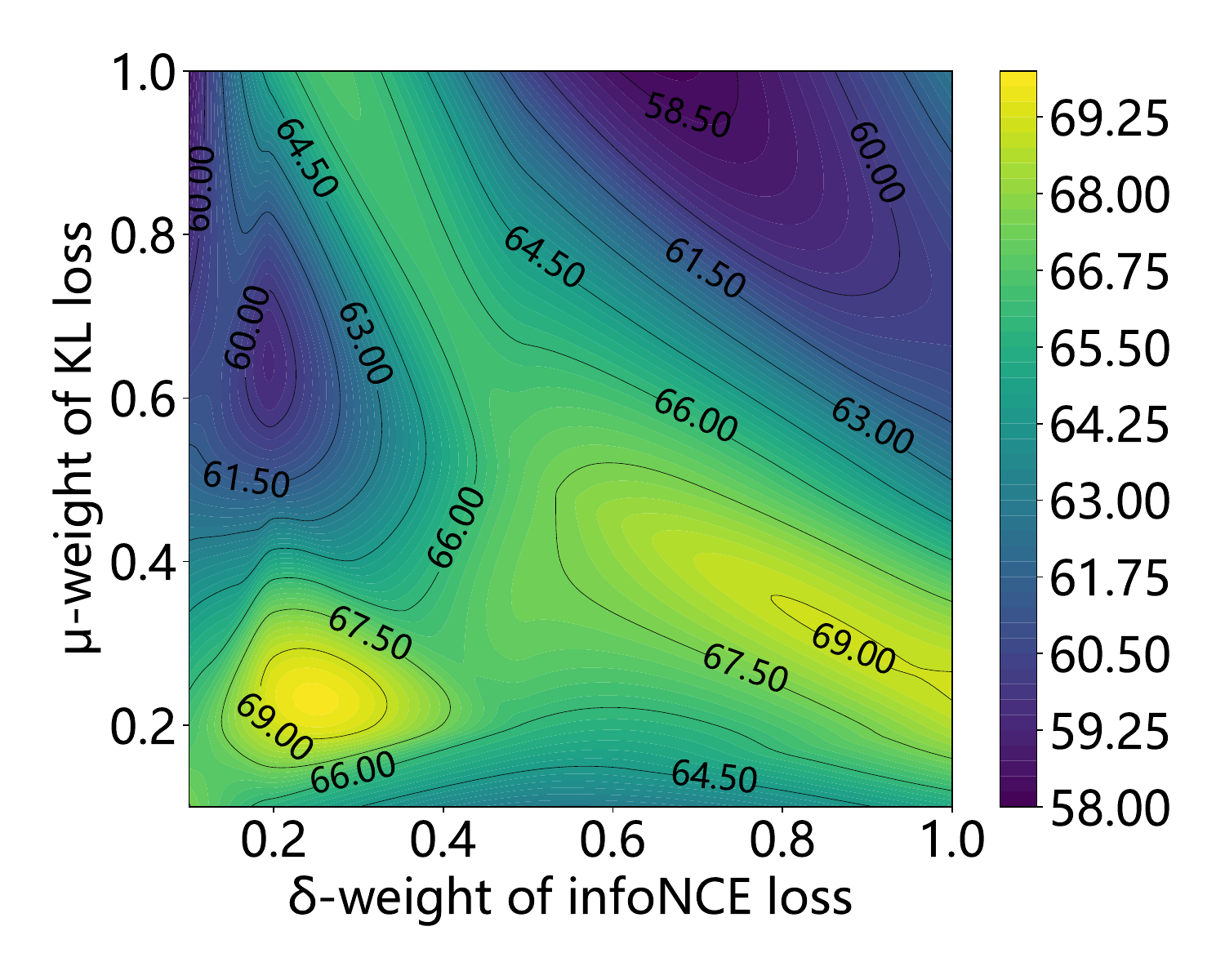}
  \end{subfigure}
  \caption{Acc. of different $\mu$ and $\delta$ under Office-10 dataset}
  \label{fig:sensitivity}
\end{figure}

% \textcolor{blue}{We compared the average accuracy of different weight combinations on the Office-10 dataset and smoothed the heat map by interpolation to observe the trend of change. As shown in Figure~\ref{fig:sensitivity}, when mu is between 0.1 and 0.4, the accuracy is high and stable, while delta is insensitive in this range.}
We conducted a hyperparameter sensitivity analysis on the Office-10 dataset by varying the loss weights $\mu$ and $\delta$ over \{0.1, 0.2, 0.5, 1.0\}. To visualize their impact, we generated an interpolated heatmap of average accuracy. As shown in Figure~\ref{fig:sensitivity}, accuracy remains high and stable when $\mu$ is within [0.1, 0.4], with $\delta$ having minimal influence in this range. Notably, the highest accuracy of 69.12 occurs at $\mu = \delta = 0.2$, followed by 68.62 at $\mu = 1.0$, $\delta = 0.2$.

The results show that our algorithm is not sensitive to the parameter changes of the loss function within a certain range and can maintain high performance.
\paragraph{Comparison of different classifier replacement methods}

\begin{figure}[htbp]
    \centering
    \begin{subfigure}{0.7\linewidth}
        \includegraphics[width=\linewidth]{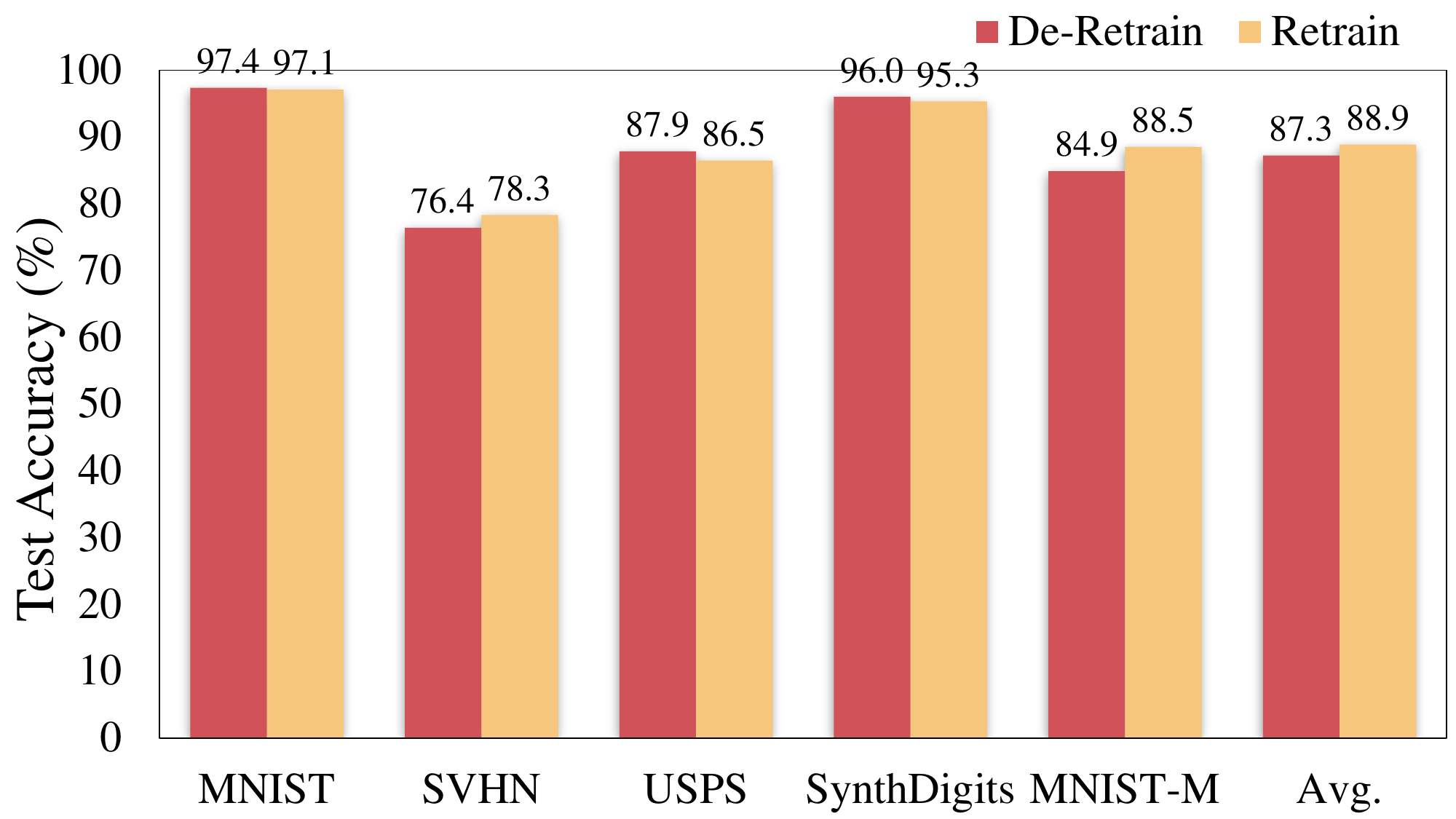}
        \caption{Digits Dataset}
        \label{fig:digits_g}
    \end{subfigure}
    \begin{subfigure}{0.49\linewidth}
        \includegraphics[width=\linewidth]{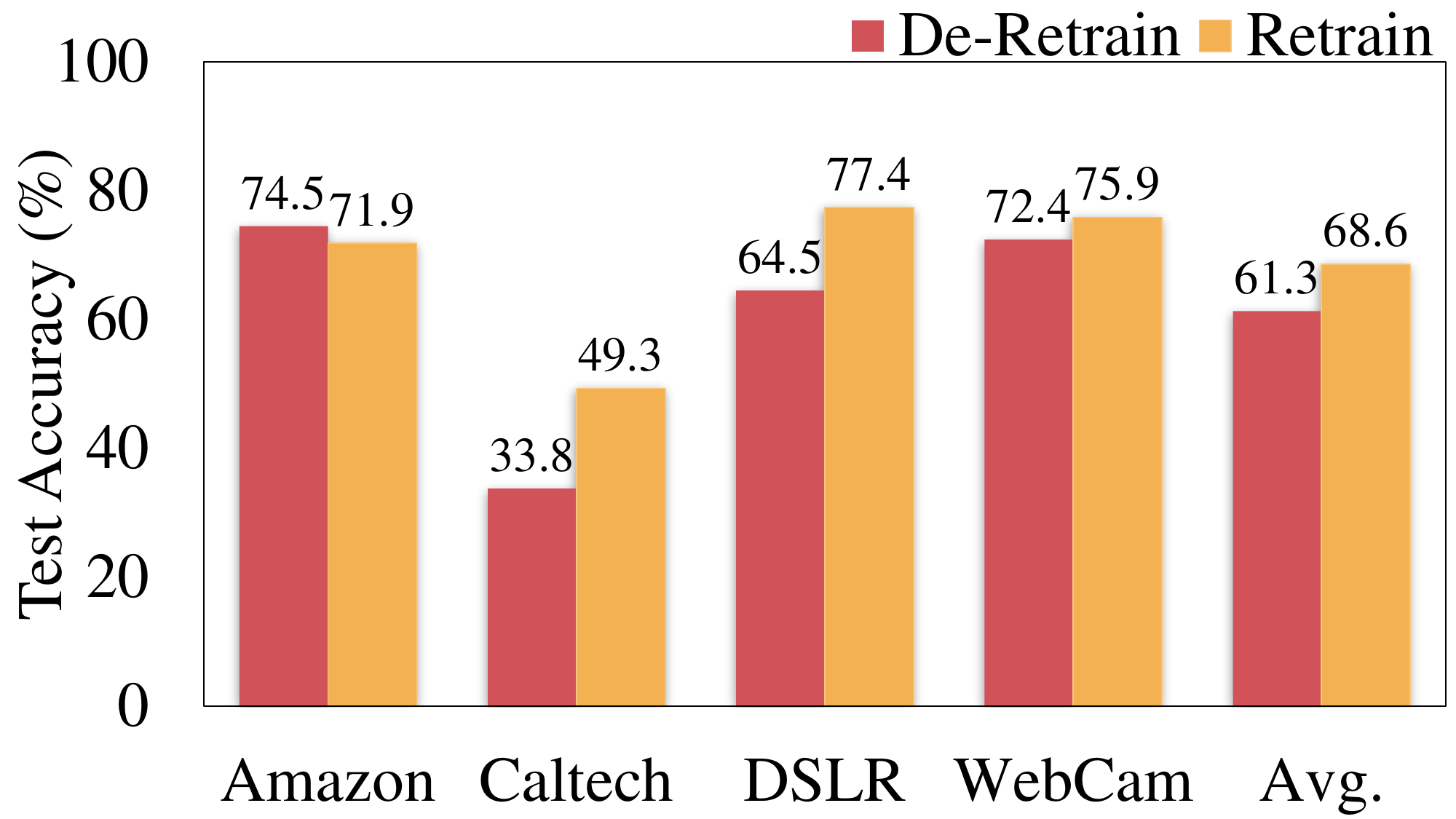}
        \caption{Office-10 Dataset}
        \label{fig:office_g}
    \end{subfigure}
    \hfill
    \begin{subfigure}{0.49\linewidth}
        \includegraphics[width=\linewidth]{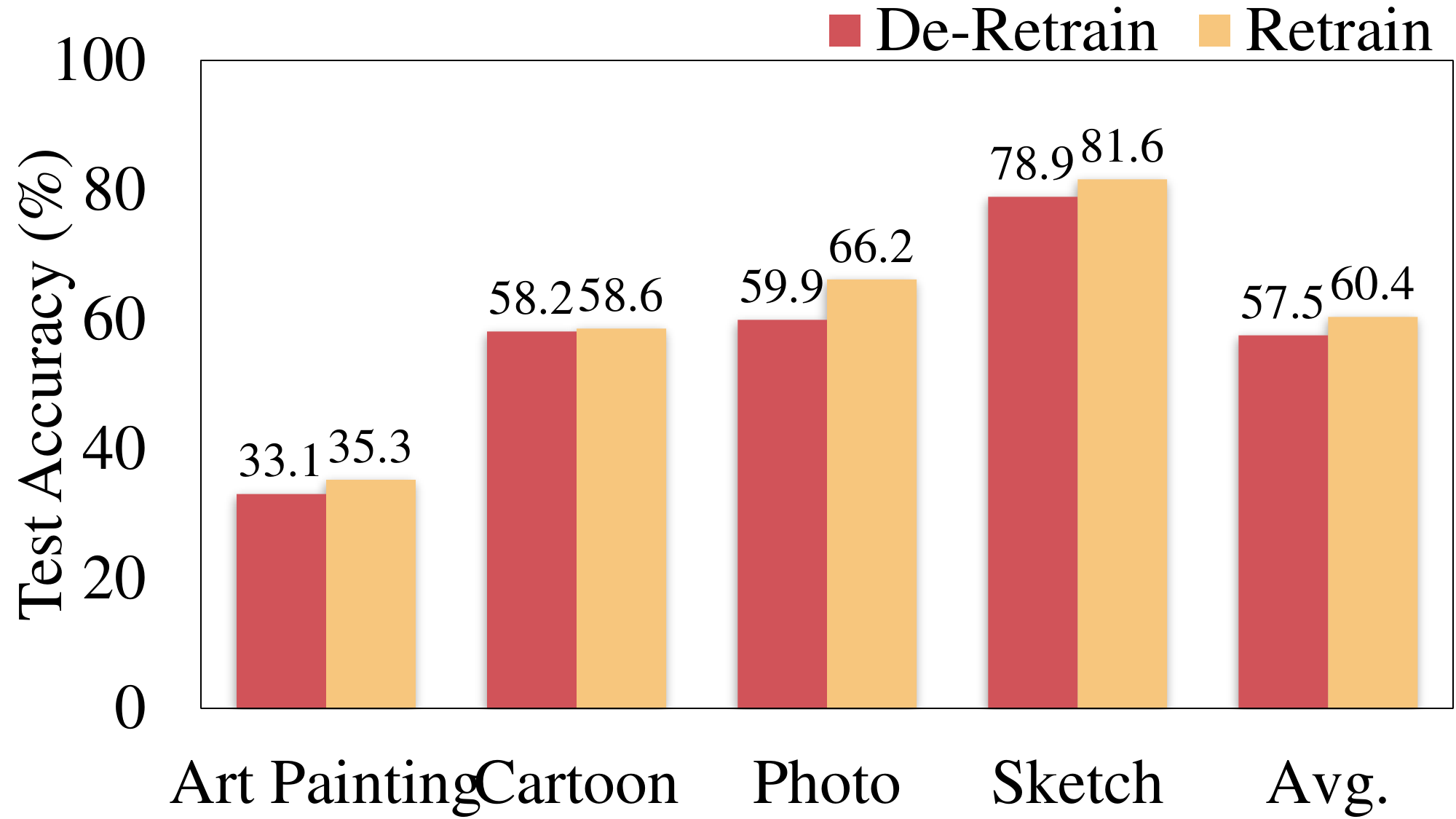}
        \caption{PACS Dataset}
        \label{fig:pacs_g}
    \end{subfigure}
    \caption{Comparison of accuracy with and without training the global classifier on the three datasets.}
    \label{fig:classifier}
\end{figure}

We evaluate the effectiveness of the global classifier through an ablation study by removing it. As shown in Figure~\ref{fig:classifier}, although the simplicity of the Digits dataset results in slightly lower accuracy for the global classifier on some sub-datasets (panel~\ref{fig:digits_g}), it still outperforms the baseline without a global classifier across multiple sub-datasets. Notably, our method achieves 3.61\% higher accuracy on MNIST-M. The benefits are more pronounced on datasets with substantial feature drift: on Office-10 (panel~\ref{fig:office_g}), the global classifier surpasses the baseline on nearly all sub-datasets, achieving 49.33\% on Caltech—a 15.55\% improvement. On PACS (panel~\ref{fig:pacs_g}), it consistently outperforms the baseline across all sub-datasets, with gains of up to 3\%.

These results confirm the necessity of FedPall’s global classifier, which captures cross-client category information to enhance client-server collaboration and improve the framework’s generalization against feature drift.

\paragraph{Privacy Leakage Risk}
% \textcolor{blue}{We use the Data Efficient Mutual Information Neural Estimator (DEMINE) to evaluate the privacy risk of prototype mixture features~\cite{lin2019data}. On the Office-10 training dataset, the mutual information (MI) values of 1) standard Gaussian noise, 2) prototype mixture only, 3) Bernoulli masking only, and 4) our features (prototype mixture + Bernoulli masking) are 2.96, 3.06, 3.04, and \textbf{2.10} (lower is better). The corresponding average accuracies are 66.38, 64.46, 63.70, and \textbf{67.55}, respectively, and the accuracy is 65.76 in the noise-free case. In addition to better privacy protection, our method also has higher accuracy because the update direction of the global prototype and feature encoder is consistent. In addition, in terms of efficiency, since the features are encrypted only after local training is completed (with linear time complexity), the computational overhead is not affected.}

We evaluate the privacy risk of prototype mixture features using the Data Efficient Mutual Information Neural Estimator (DEMINE)~\cite{lin2019data}. On the Office-10 training set, the mutual information (MI) scores for: (1) standard Gaussian noise, (2) prototype mixture only, (3) Bernoulli masking only, and (4) our method (prototype mixture + Bernoulli masking) are 2.96, 3.06, 3.04, and \textbf{2.10}, respectively (lower is better). The corresponding average accuracies are 66.38, 64.46, 63.70, and \textbf{67.55}, compared to 65.76 in the noise-free case. Our approach not only offers stronger privacy protection but also improves accuracy, attributed to the consistent update direction between the global prototype and feature encoder. Additionally, since encryption is applied post-training with linear time complexity, the computational overhead remains negligible.

%% file: sec/6_conclusion.tex
\section{Conclusion}
In this study, we focus on the feature drift problem in FL. The feature drift problem causes the same class samples on different clients to have distinct feature distributions, making it difficult for traditional model aggregation methods to handle such data heterogeneity.
To tackle this problem, we design a prototype-based adversarial collaborative framework to unify feature spaces and enhance classification boundaries. The global classifier is retrained with mixed features to further grasp classification-relevant information from a global perspective. Our method has empirically achieved state-of-the-art performance in popular feature-drifted datasets with multiple data sources.

% Currently, our work is only designed for classification tasks and evaluated with image-based datasets.In the future, we aim to systematically validate the framework's generalizability in more task types with data of other modals.

\section*{Acknowledgements}
This work was supported in part by the Innovation Team Project of Guangdong Province of China (2024KCXTD017) and Guangdong-Hong Kong-Macau University Joint Laboratory (2023LSYS005).
%Our work can provide some insights for future research on solving feature drift issues.
% \bibitem{b1} G. Eason, B. Noble, and I. N. Sneddon, ``On certain integrals of Lipschitz-Hankel type involving products of Bessel functions,'' Phil. Trans. Roy. Soc. London, vol. A247, pp. 529--551, April 1955.
% \bibitem{b2} J. Clerk Maxwell, A Treatise on Electricity and Magnetism, 3rd ed., vol. 2. Oxford: Clarendon, 1892, pp.68--73.
% \bibitem{b3} I. S. Jacobs and C. P. Bean, ``Fine particles, thin films and exchange anisotropy,'' in Magnetism, vol. III, G. T. Rado and H. Suhl, Eds. New York: Academic, 1963, pp. 271--350.
% \bibitem{b4} K. Elissa, ``Title of paper if known,'' unpublished.
% \bibitem{b5} R. Nicole, ``Title of paper with only first word capitalized,'' J. Name Stand. Abbrev., in press.
% \bibitem{b6} Y. Yorozu, M. Hirano, K. Oka, and Y. Tagawa, ``Electron spectroscopy studies on magneto-optical media and plastic substrate interface,'' IEEE Transl. J. Magn. Japan, vol. 2, pp. 740--741, August 1987 [Digests 9th Annual Conf. Magnetics Japan, p. 301, 1982].
% \bibitem{b7} M. Young, The Technical Writer's Handbook. Mill Valley, CA: University Science, 1989.
% \end{thebibliography}